\documentclass{article}
\usepackage[utf8]{inputenc}
\usepackage{hyperref}
\usepackage[backend=bibtex, sorting=none]{biblatex}
\usepackage{graphicx}
\usepackage{wrapfig}
\usepackage{float}
\usepackage{tabularx}
\usepackage[parfill]{parskip}
\usepackage{authblk}
\author{Huahong Tu, Siyuan Peng, Vladimir Leung, Richard Gao \\
University Of Maryland, College Park, MD 20740\\
{{\{hh2, peng2000, vlade813, rgao1\}@umd.edu}}}
\date{}
\title{SoK: Vehicle Orientation Representations for Deep Rotation Estimation}

\begin{document}

\maketitle

\begin{abstract}
%Why did you do the study?
%What did you do?
%What did you find? and…
%What did you conclude? What’s the significance of the study?

In recent years, there is an influx of deep learning models for 3D vehicle object detection. However, little attention was paid to orientation prediction. Existing research work proposed various vehicle orientation representation methods for deep learning, however a holistic, systematic review has not been conducted. Through our experiments, we categorize and compare the accuracy performance of various existing orientation representations using the KITTI 3D object detection dataset \cite{kitti_geiger_2012}, and propose a new form of orientation representation: Tricosine. Among these, the 2D Cartesian-based representation [cos($\theta$), sin($\theta$)], or Single Bin, achieves the highest accuracy, with additional channeled inputs (positional encoding and depth map) not boosting prediction performance. Our code is published on GitHub: \href{https://github.com/umd-fire-coml/KITTI-orientation-learning} {https://github.com/umd-fire-coml/KITTI-orientation-learning}
\end{abstract}

\section{Introduction} \label{section:introduction}
% What is the motivation for representing rotation differently other than radians?
Within vehicle object detection tasks, rotation estimation typically receives little attention, due to the main tasks, object localization and dimension estimation, receiving more focus. We believe that finding ways to improve orientation estimation is an important task, because accurate orientation estimation is useful for dependent tasks such as path projection \cite{along2006pathpoject,geiger2014platforms,alvares2014roaddetection},
birds-eye-view visualization \cite{sindagi2019mvx,veddersparse,saha2021enabling,roddick2021learning}, and dimensions estimation \cite{camastra2016intrinsic,li2009improved, campadelli2015intrinsic}.

%, joint prediction \cite{chen2017multi,ku2018joint,engelcke2017vote3deep,wang2020pillar},

However, accurate rotation estimation contains challenges that seems to be highly dependant on the rotation representation target for deep learning. With the default angle representations in radians/degrees (based on either global or local camera angle), the problem of \textit{discontinuity} occurs when rotation values $\theta \in [-\pi,\pi]$ circularly jump from $-\pi$ to $\pi$, potentially resulting in huge losses for small angle changes across the wrap around boundary.

Some existing literature aim to solve this problem by mapping $\theta$ in a cosine or sine function, that can provide a circular, continuous representation while maintaining the compactness of one output value. However, using this alone creates an \textit{ambiguity} problem, where cos($\theta$) and sin($\theta$) values now map to two $\theta$ values within [-$\pi$,$\pi$]. 

Discontinuity and ambiguity can be solved with a 2-D Cartesian representation using (cos($\theta$), sin($\theta$)) pair, creating a continuous, unambiguous representation \cite{continuity_zhou_2020} for every $\theta \in [-\pi,\pi]$. Some representations further this approach, treating this pair as a sub-component of a larger representation by dividing the [-$\pi$,$\pi$] range into equal sized "bins" and isolating the orientation representation to a fixed  range~\cite{voting_bin_zhao_2020}. A research paper further extends this multiple bins representation by adding a component representing a confidence value on top of the (cos($\theta$), sin($\theta$)) pair values within each bin \cite{multibin_mousavian_2017}. 

% Why we still don't know what is the most 'accurate" way of representing rotation?

However in 2021, much of the existing Vehicle Object Detection research selects one of various rotation representations for prediction without much cited test results or numerical evidence for their choice. Current literature has more than 5 ways to represent a vehicle's rotation without a clear unified message of which is the best way for representation, and whether switching to different representation could affect prediction accuracy.

% Since we don't know, What is the correct way to do experiments to find out?
Hence, our work aims to provide numerical results for the most accurate type of vehicle orientation representation, in hope that future object detection work will cite this evidence. We implemented various type of orientation representations and compared their results. For every type of orientation representation, we used a similar backbone and training setup, with the only variable being the final output layer to represent the vector that corresponds to the component values for the type of orientation representation.

We also extend our analysis with additional inputs such depth map and positional encoding, to see if providing depth and localization information results in improved performance respectively.

This process can be easily replicated to more complicated orientation representations with more rotational axes, other inputs like LiDAR or stereo images, and other datasets.

With this systematization of knowledge paper we make the following contributions:

\begin{itemize}
    \item Develop a process for testing various types of vehicle orientation representation.
    \item Categorize and clarify various commonly used orientation representations.
    \item Train and evaluate different types of representations for accuracy and complexity.
    \item Discuss why among these, 2-D Cartesian representation is the best type of representation for deep rotation estimation.
\end{itemize}

\section{Vehicle Orientation Representations}
\begin{figure}[!h]
    \centering
    \includegraphics[width=120mm]{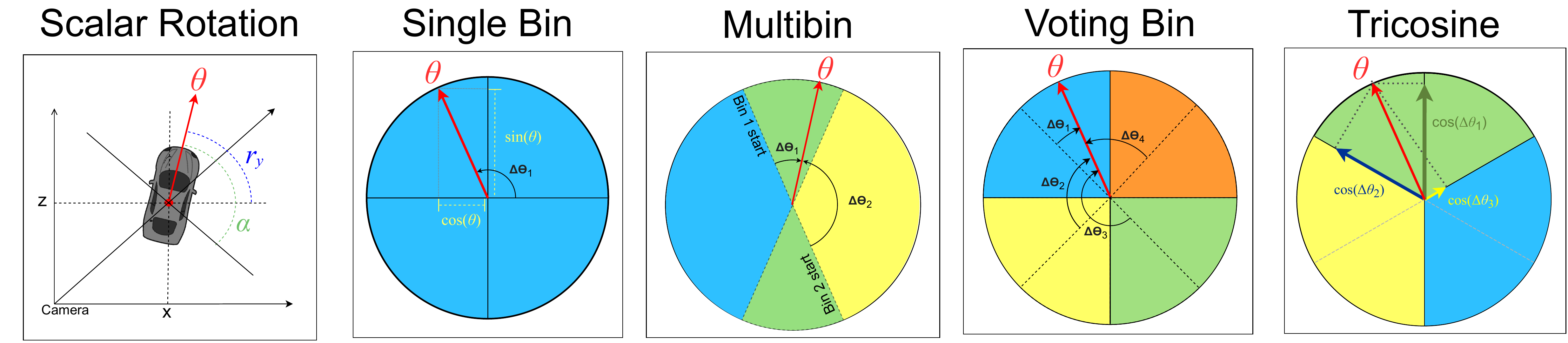}
    \caption{A visualization of the presented orientation methods}
    \label{fig:all_orien}
\end{figure}

\subsection{1-D Scalar Representations}
Orientation representations in this section measure an object's orientation as the rotational offset between the angle and a camera reference axis, in radians or degrees. These representations benefit from compactness, representing orientation with a single scalar value, which can reduce the model's final output layer to a single node. However, Grassia~\cite{grassia1998practical} states that Euler angles and quaternions representation\cite{Simonelli_2019_ICCV} are not ideal to compute and differentiate positions, due to the \textit{discontinuity} \cite{ming2021optimization} and \textit{ambiguity} problem. We still observe recent researchers such as \cite{cai2020monocular,chabot2017deep,chen2021monorun, wang2021depth, park2021pseudo,NIPS2015_6da37dd3,Shi_2021_ICCV,Luo_2021_CVPR,Ding_2020_CVPR, Ma_2019_ICCV,wang2021detr3d, lin2021single,Liang_2019_CVPR} using these types of scalar-based angle representation in 3D Object Detection. We are also noticing some variants of the scalar-based methods. Wang et al. \cite{wang2021fcos3d} claimed better performance when dividing the angles into 2-bins with each bin represented by angle $\theta$ (offset) with period $\pi$ to distinguish two opposite orientations, but such method suffer from the \textit{discontinuity} and \textit{ambiguity} problem for values at the border of each bin. Shrivastava and Chakravarty \cite{shrivastava2020cubifae} normalized orientation angle to (0,1). Carrillo and Waslander \cite{carrillo2021urbannet}, Gahlert et al \cite{gahlert2020single}, and Plaut et al \cite{plaut20213d} studied the multiple degrees of freedom rotation of vehicle by applying regressing on scalar Euler angle independently on each axis.

However, all 1-D scalar-based orientation representation suffer from non-continuity \cite{continuity_zhou_2020}, where model predictions near the boundary of the orientation's range can result in an extreme loss even if the  angular prediction is close to the ground truth, e.g. a 2$\pi$ rad prediction and 0 rad ground truth incurs an extremely large loss for the prediction, when using L1/L2 loss.

Some previous works attempts to alleviate the non-continuity problem \cite{single_bin_hara_2017} by having a customized loss function based on the 2D angular difference instead of the L1/L2 distance on the scalar value. However, these loss functions are not converging between prediction and ground truth angles that are exactly $\pi$ radians apart since they have near-zero gradient values. 

It also needs additional considerations to ensure that it penalizes out-of-bound scalar prediction values, such as for values below $-\pi$ or above $\pi$. The loss function should not simply use 2D angular difference for loss as it would be able to correct extreme scalar prediction values.

A customized loss function is also prone to implementation errors, as it needs to be compatible with the automatic differentiation engine of machine learning frameworks (e.g. tensorflow and pytorch). It also require re-implementation and testing for each specific framework, unlike common loss functions such as L1/L2 loss. 

\subsubsection{Rotation-Y (Global/Egocentric Rotation)}
% Motivation for Rotation Y, i.e why this is better than $\alpha$
\begin{figure}[!h]
    \centering
    \includegraphics[width=50mm]{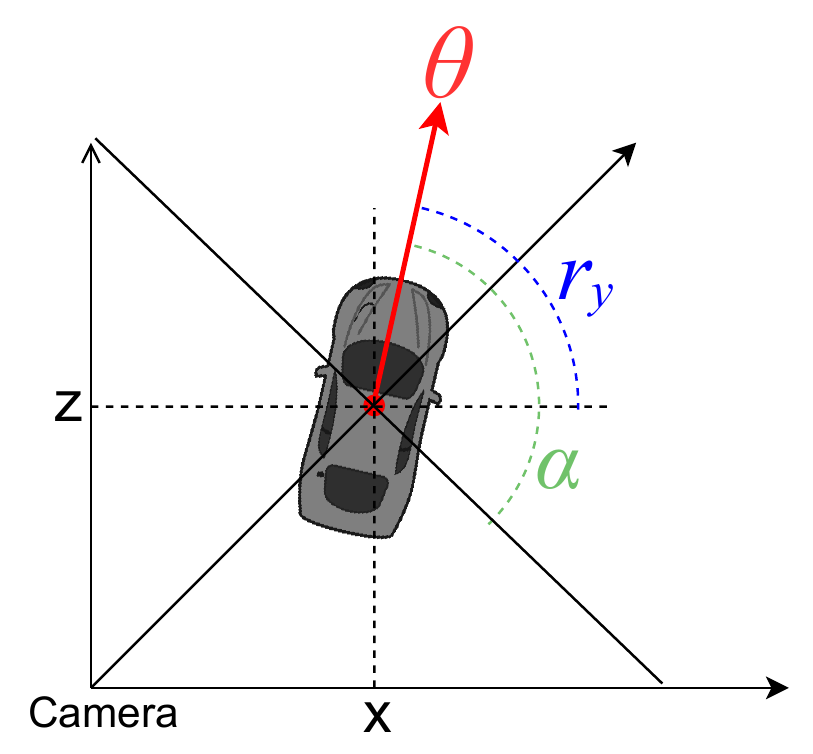}
    \caption{The $r_y$ and $\alpha$ radial observation angle forms given by KITTI. The $\alpha$ representation contains the car's appearance to the camera, while $r_y$ depicts the vehicle relative to the scene.}
    \label{fig:Rotation_radians}
\end{figure}
\begin{figure}[!h]
    \centering
    \includegraphics[width=100mm]{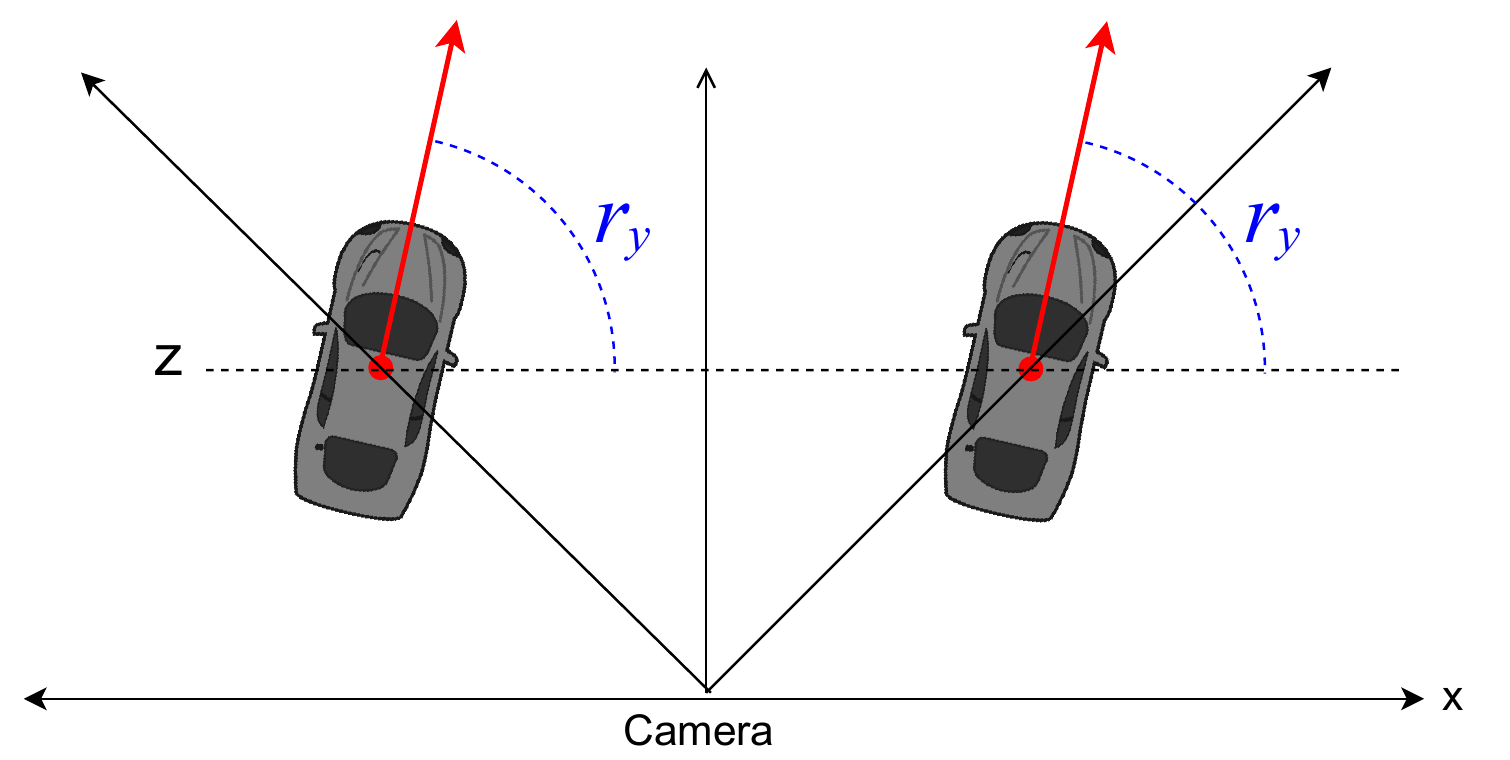}
    \caption{Example of radial angle form Rotation Y held constant. The vehicle's orientation now does not depend on what the camera sees, but rather the orientation of the vehicle within the scene.}
    \label{fig:Rotation_y_example}
\end{figure}

Rotation-Y, $r_{y}$, also know as camera z-axis rotation or egocentric global rotation \cite{kitti_geiger_2012,deng2021revisiting}, is one of two orientation representations given in KITTI object detection labels~\cite{kitti_geiger_2012}. It measures the object's orientation in radians ($r_{y}$ $\in [-\pi,\pi]$), relative to the axis perpendicular to the camera's forward facing z-axis. Two objects with the same $r_{y}$ value will point in the same direction relative to the scene regardless of the camera location. See Figure \ref{fig:Rotation_y_example} for changes that can occur with constant global rotation. 

\subsubsection{Alpha (Local/Allocentric Rotation)}
\begin{figure}[!h]
    \centering
    \includegraphics[width=100mm]{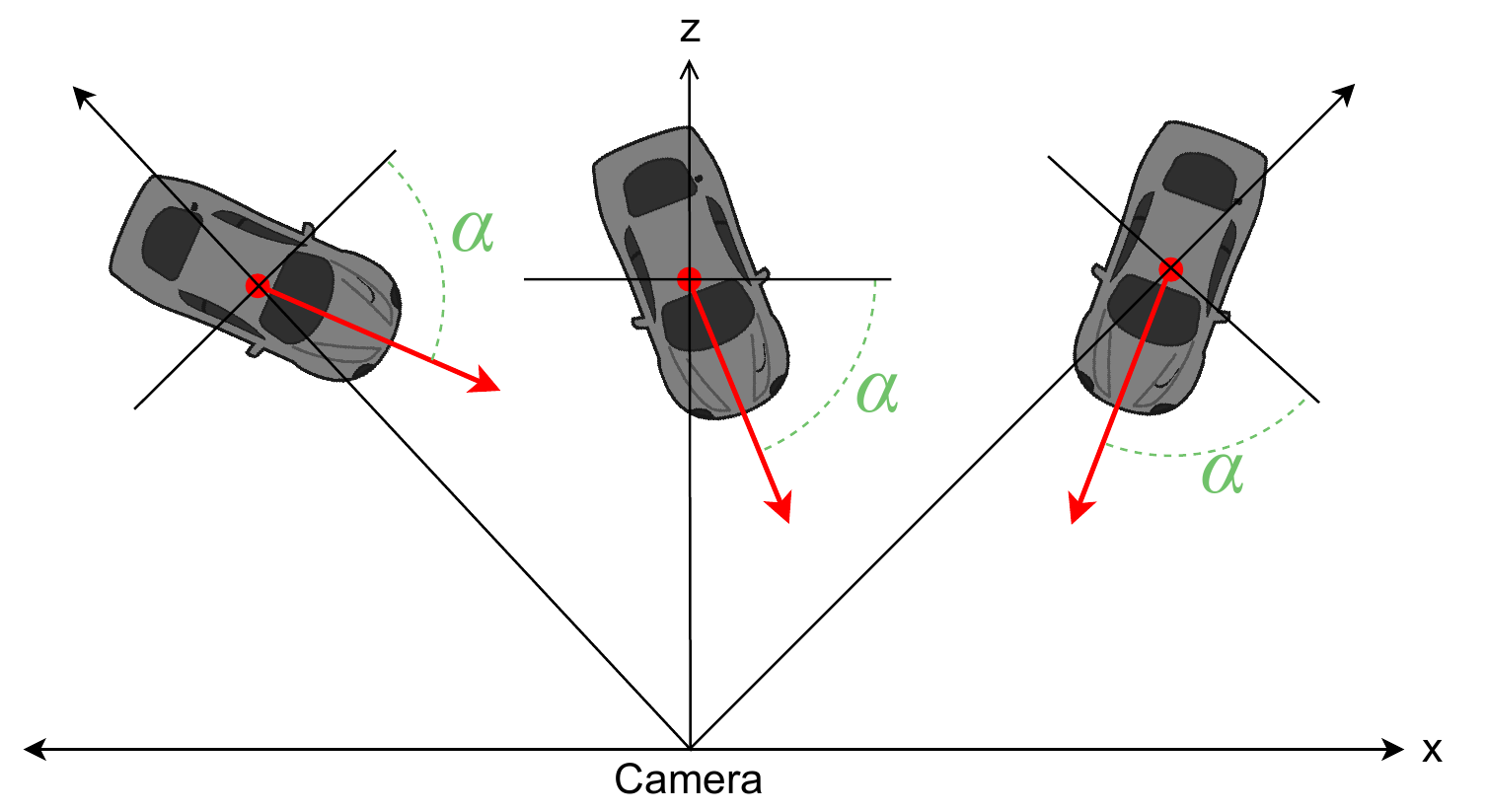}
    \caption{Example of radial angle form $\alpha$ held constant. Observe that the vehicle will have the same appearance to the camera from the shown locations.}
    \label{fig:Alpha_example}
\end{figure}

Alpha, $\alpha$, also known as the local observation angle of the object, is another scalar orientation representation provided in the KITTI object detection labels \cite{kitti_geiger_2012}. It measures the object's orientation in radians ($\alpha \in [-\pi,\pi]$),  relative to camera's observation angle towards the center of the object. Figure \ref{fig:Rotation_radians} provides a bird's eye view visual representation of how $\alpha$ is measured. See Figure \ref{fig:Alpha_example} for visual examples of changes that can occur with constant local rotation. 

Two objects with the same $\alpha$ value will have a similar visual appearances \cite{liu_deep_2019}, e.g. car headlights directly facing the camera (see Figure \ref{fig:Alpha_example}). Therefore, it may be easier for a model to learn to predict the $\alpha$, given the camera local appearance of an object. However, because camera sensors are usually flat (not curved to keep a constant focal distance) \cite{zhou2020flat}, optical distortions at different frame positions, e.g. an object near the edge of the frame vs an object at the center of the frame, can produce varying camera observed distorted appearances despite having same $\alpha$ values.

To draw and visualize a object's orientation from a bird's eye view perspective, the object's $\alpha$ will need to be converted into $r_{y}$. With the object's x and z location information, we can convert $r_y$ to $\alpha$ or vice versa \cite{li2021sm3d,you2019pseudo,xu2020zoomnet}, using the formula:
\begin{equation}
r_{y} = \alpha + \arctan \frac{x}{z}
\end{equation}
where $x$ represents the object's horizontal position, and $z$ represents the object's forward distance, both relative to the center of the camera. 

%\subsection{Discussion}
\subsection{Single Bin (2-D Cartesian Representations)}
\begin{figure}[H]
    \centering
    \includegraphics[width=50mm]{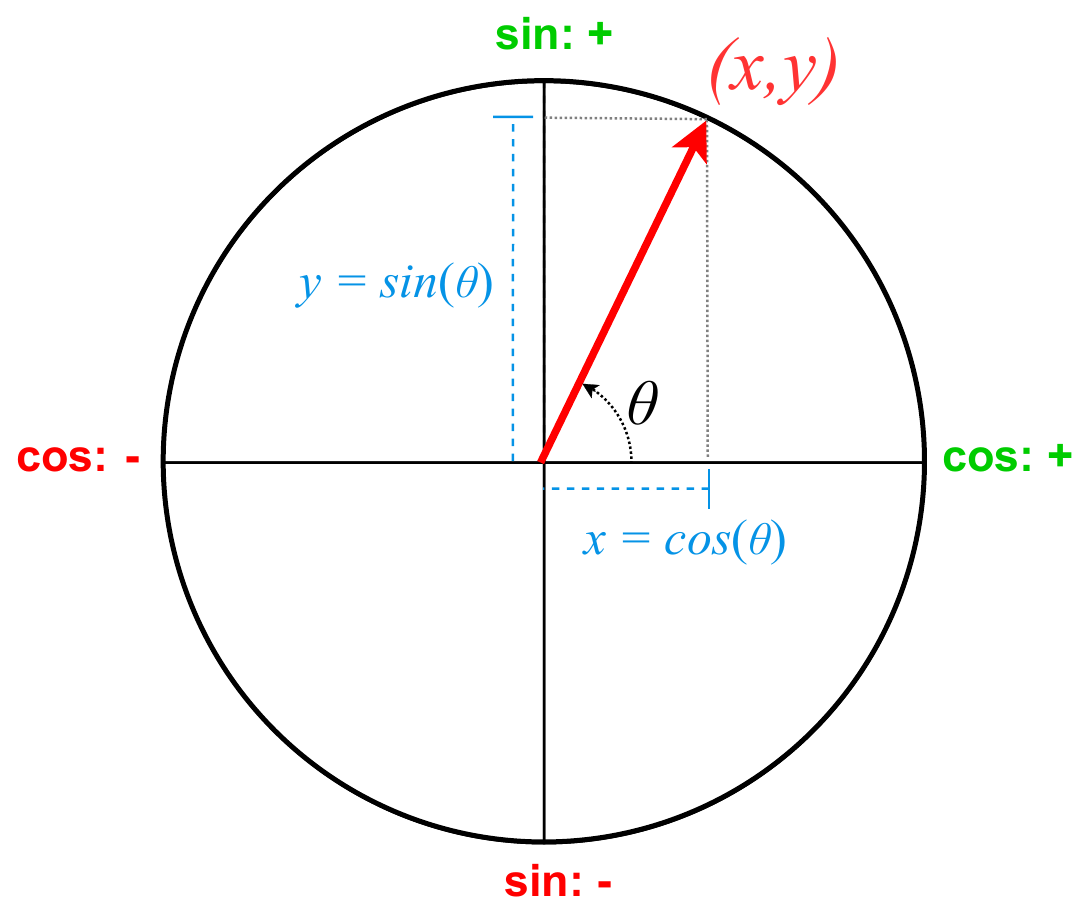}
    \caption{A reference displaying the mapping from angles to Cartesian coordinates}
    \label{fig:Angle_Cartesian_Reference}
\end{figure}

The discontinuity problem can be partly resolved by representing the angle using the cosine of the angle \cite{he2020svga, zheng2021se, reading2021categorical, lang2019pointpillars,gao2020monocular} or sine of the angle \cite{rukhovich2021imvoxelnet,nobis2021radar} only due to its periodic properties. However, only using one of the cosine and sine of the angle cannot be unambiguously converted back into the angle without using at least 3-bins. Furthermore, when the angular difference is $\delta = \pm\pi$, the differentiation results of such angle, i.e. $cos(\delta)$, are zero, which could stop backpropagation learning at the extreme opposite output. To resolve this problem, researchers \cite{Liu_2020_CVPR_Workshops,deng2020voxel,liu2021ground,shi2021geometry,Julca-Aguilar_2021_ICCV,Chai_2021_CVPR, jorgensen2019monocular, naiden2019shift,Yin_2021_CVPR,yang2018pixor,roddick2018orthographic,yang2018hdnet} performed regression on both the cosine and sine of the angle (cos($\theta$), sin($\theta$)). When converting any given 1D Scalar-Based Orientation $\theta$ into 2-Dimensional values, they can be mapped to 2-D Cartesian system with angle's cosine value projected to x-axis and sine value projected to y-axis. \cite{te2020relations}
Figure \ref{fig:Angle_Cartesian_Reference} provides a visual representation of how the 2D Cartesian representation of an object's orientation is measured. 

2D Cartesian-based representation can be unambiguously converted back into $\theta$ in radians using the formula:
\begin{equation}
\theta = \arctan \frac{cos(\theta)}{sin(\theta)}
\end{equation}

This type of angle representation avoids the non-continuous boundary problem in 1-D Scalar-Based methods, allowing a model to use L1/L2 loss to penalize on the angular difference while also penalizing out-of-bound predictions, creating a smooth gradient for convergence, unlike singular value representations. However, even with L1/L2 loss during training, model predictions can still make out-of-bound values, so additional post-processing is needed to clip or scale the predicted values within the valid cosine and sine ranges.

\subsection{N-Bin and Affinity Representations}
The discontinuity issue can also be resolved by converting the angle $\theta$ into a discrete classification problem, by using multiple confidence values to represent the affinity for each bin. This generalizes to any other methods where the highest value first determines what bin the angle lies in. We dub these as \textbf{Affinity} based representations. Liao et al. \cite{liao2019spherical} stated that converting the angle to discrete classification forms lead to stable training and convergence. Using discrete bins has a potential trade-off between the number of bins that leads to longer training time due higher number of final layer nodes and the prediction accuracy, so final angle prediction accuracy depends on the number of bins and specific implementations\cite{liao2019spherical}. Researchers \cite{shi2019pointrcnn, sun2021rsn, Chen_2016_CVPR} first predict on a few discrete bins and then regress on the offset. There are also works designed to utilize a large number of bins or more exotic techniques: 12 non-overlapping equal bins \cite{ma2021delving}, 72 non-overlapping bins \cite{ barabanau2019monocular}, mean-shift after binning twice \cite{single_bin_hara_2017}.

% All of the representation methods in this section measure the orientation of an object based on a list of affinity values in N discrete evenly divided angular bins. Given an object's direction, we desire a higher affinity value for bins with a lower angular distance (being closer) to the object's orientation.

However, it is important to reduce the number of parameter in representing the rotation as the number of parameters directly affect the number of node in the final output layers. There are also complications in combining loss functions, as it creates loss competitions and non-smooth loss gradients, resulting in convergence issues during training. For these reasons, we selected to test the most commonly used techniques.

% this one too for pedestrian pose estimation: https://arxiv.org/pdf/1907.06777.pdf
% \begin{wrapfigure}{h}{30mm}
%     \centering
%     \includegraphics[width=30mm]{diagrams/Single_Bin.pdf}
%     \caption{}
%     \label{fig:Single_Bin}
% \end{wrapfigure}
% \newpage

\subsubsection{Multibin \cite{multibin_mousavian_2017}}

\begin{figure}[H]
    \centering
    \includegraphics[width=80mm]{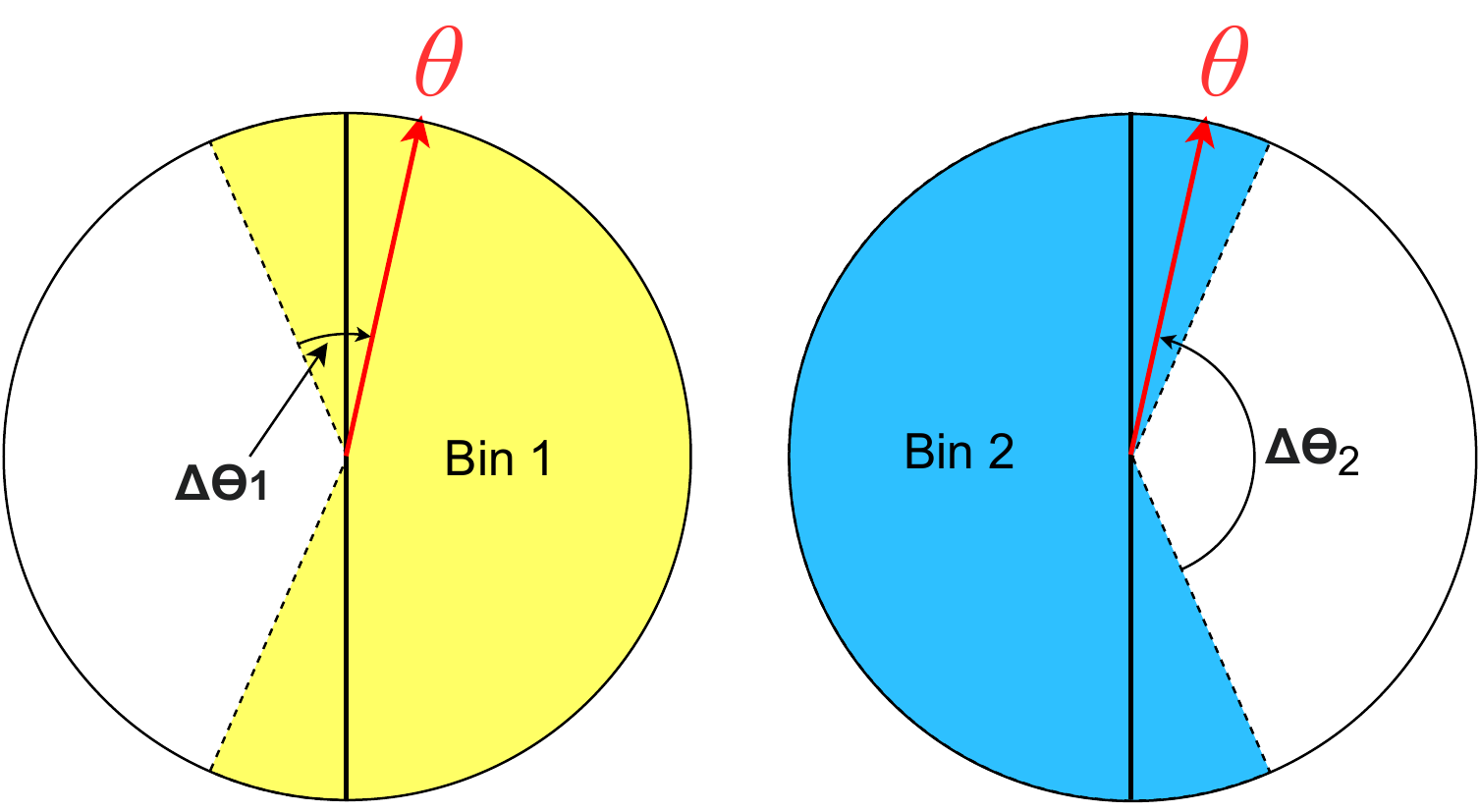}
    \caption{Multibin based representation. Each bin is encoded with a 2-D Cartesian value pair (or Single Bin) and a confidence value. Offset and non-offset cases provide differing bin/confidence values to create a more continuous form \cite{multibin_mousavian_2017}.}
    \label{fig:multibin}
\end{figure}

 Multibin attempts to approach orientation estimation by classifying a bin with the highest confidence values and overlapping bins. Mousavian et al.\cite{multibin_mousavian_2017} poses that the model should predict a confidence vector to first classify one bin, then reverse this bin's sine, cosine value pair. The authors have outstanding performance with only two bins, utilizing a 10\% total overlap at the bins' borders, see Fig.\ref{fig:multibin}. This overlap assists with identifying angles close to the edge of these bins, which is elaborated upon later. Multiple papers \cite{chen2020monopair, zhou2019objects, zhang2021learning, ku2019monocular,gs3d_li_2019,liu2019deep,lu2021geometry,Zhou_2021_CVPR,Hu_2019_ICCV,liu2021autoshape,qi2018frustum,wang2021progressive, peng2021ocm3d, li2021monocular, bao2020object, li2019multi,ku2019monocular} utilize this technique. The diagram \ref{fig:multibin} shows the angle, from each bin start to the angle. 
 
 \begin{figure}[h!]
        \centering
        \includegraphics[width=80mm]{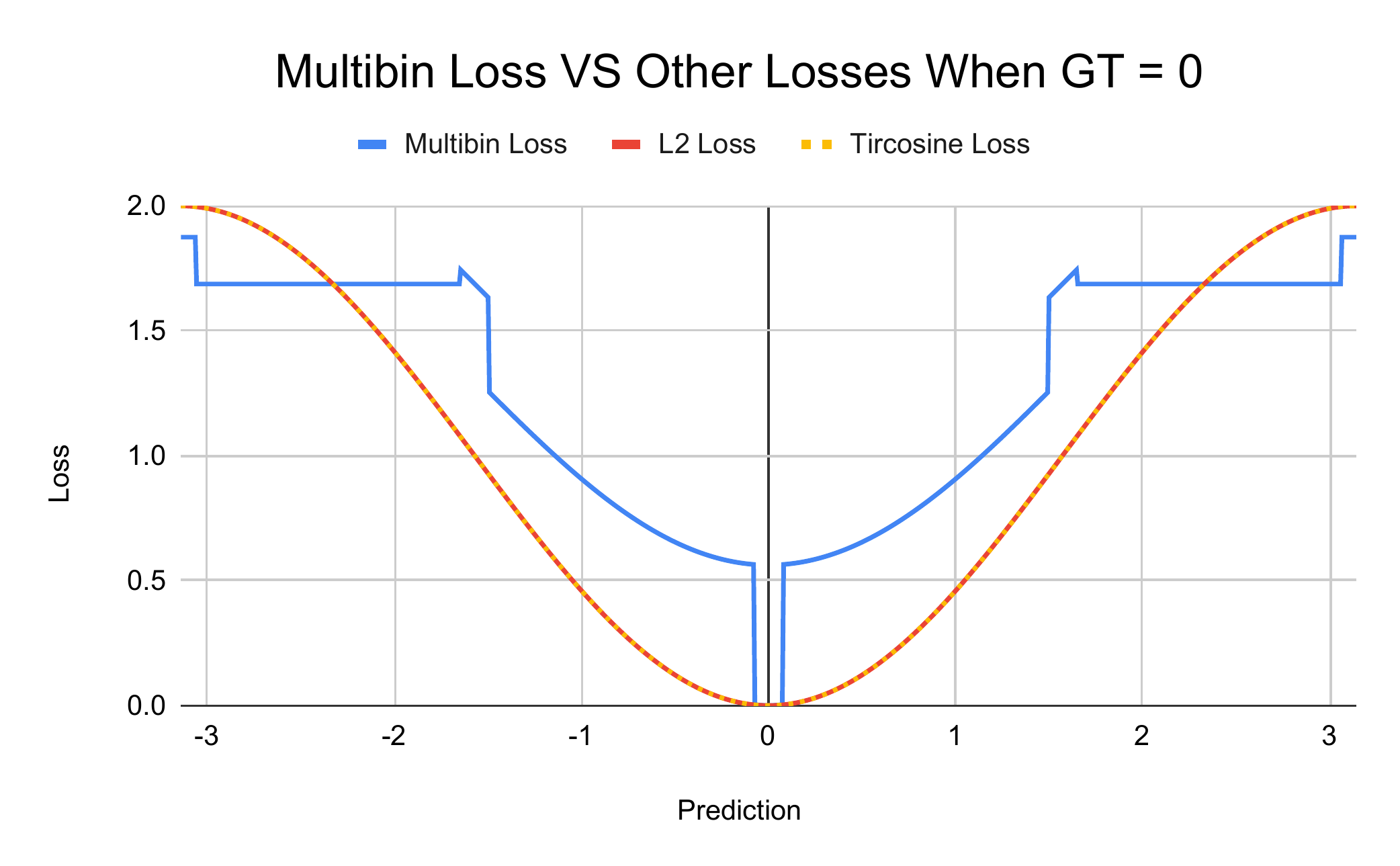}
        \caption{Graph of loss value for Multibin Loss, L2 losses and Angular Loss when the ground truth is set to zero radians. 
         i.e. let $\theta_{gt}$ be the ground truth, $\theta_{pred}$ be the predicted value and $loss$ be the loss function. For all $\theta_{pred} \in [-\pi, \pi]$ hypothetical predicted outputs, we convert them into their corresponding orientation representations using the conversion algorithm $C$, and then plot $loss(C(\theta_{pred}),C(\theta_{gt}=0))$.
        }
        \label{fig:loss_function}
    \end{figure}

In Figure \ref{fig:loss_function}, we explore the smoothness of different loss functions. Angular loss overlaps with the l2 loss. For multibin~\cite{multibin_mousavian_2017}, although the graph indicates lowest loss at zero, multibin does not have a smooth loss function. When the predicted value is within around $[\frac{1}{2}\pi,\pi]$, the gradient of loss function is zero, which gives no feedback to the model and traps them in local minimum. 

To signal which bin the angle $\theta$ lies, it assigns a confidence value to each bin's Cartesian pair values. Labels are given confidence values as follows. A confidence value of 1.0 means $\theta$ lies in that bin and not within the overlap region, while 0.0 confidence means $\theta$ does not lie in that bin. For $\theta$ within the overlapping region, a confidence of 0.5 is given to each overlapping bin. 

Multibin also labels each bin with cos($\Delta\theta$) and sin($\Delta\theta$), where $\Delta\theta$ is the angular distance to the bin's starting edge, i.e. offset. For the overlapping case, when $\theta$ is inside any overlapping region, both bins are given the cosine and sine values of each bin's respective offset. For the non-overlapping case, the bin containing $\theta$ receives the cosine and sine of the offset, while the other bin is given all zeros. 

%The offset is obtained by subtracting $\alpha$ from the starting edge angle of the current bin. If $\alpha$ is outside the total overlap range of a bin, the bin values are zeros.

Hence, Multibin attempts to use both the 2-D Cartesian and N-D affinity values in the representing the final orientation. However, this approach make it difficult to use a simple L1/L2 loss. A custom loss function will be needed to take into 0/1 confident predictions and 0.5/0.5 confident affinity predictions for overlapping bins. Additionally, we have found overlap prevents convergence, so we remove it and call this simpler form Confidence Bin. 

\subsubsection{Voting Bins}

\begin{figure}[H]
    \centering
    \includegraphics[width=50mm]{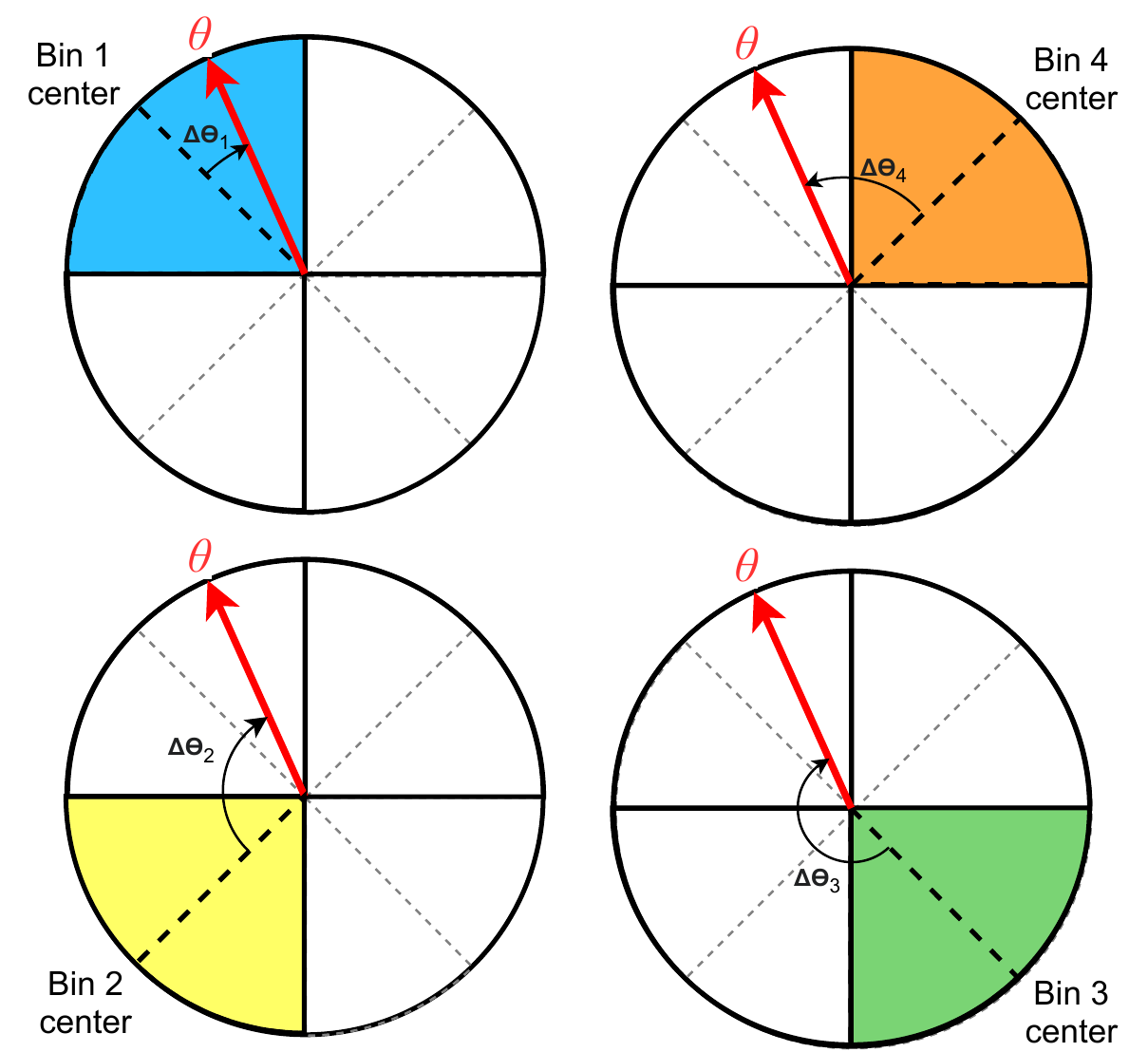}
    \caption{Voting bin based representation. Each bin is encoded with a 2-D Cartesian value pair (or Single Bin). Modeled after Multi-bin without confidence, the bins' values are averaged after the voting/exclusion algorithm.}
    \label{fig:voting_bin}
\end{figure}

Zhao et al \cite{voting_bin_zhao_2020} proposes FFNet requiring 3 or more bins, instead of explicitly using confidence values for averaging, to unambiguously convert it back into a 1-D representation.
Each bin's values encode the angular offset $\Delta\theta$ from the middle of each bin (instead from the start of each bin). The original author recommends 4 bins. Converting back to a 1-D representation is an average of all candidate bin's reversed offset angles, where any heavily outlying bins are "voted" out or excluded from the average by the algorithm below:

If any bin meets both conditions below, they are voted out of the averaging.
\begin{itemize}
    \item It and all other bins' difference are below a fixed threshold (30 degrees)
    \item All other bins' difference are also below the threshold (dynamic)
\end{itemize}

\subsubsection{Tricosine}

\begin{figure}[H]
    \centering
    \includegraphics[width=60mm]{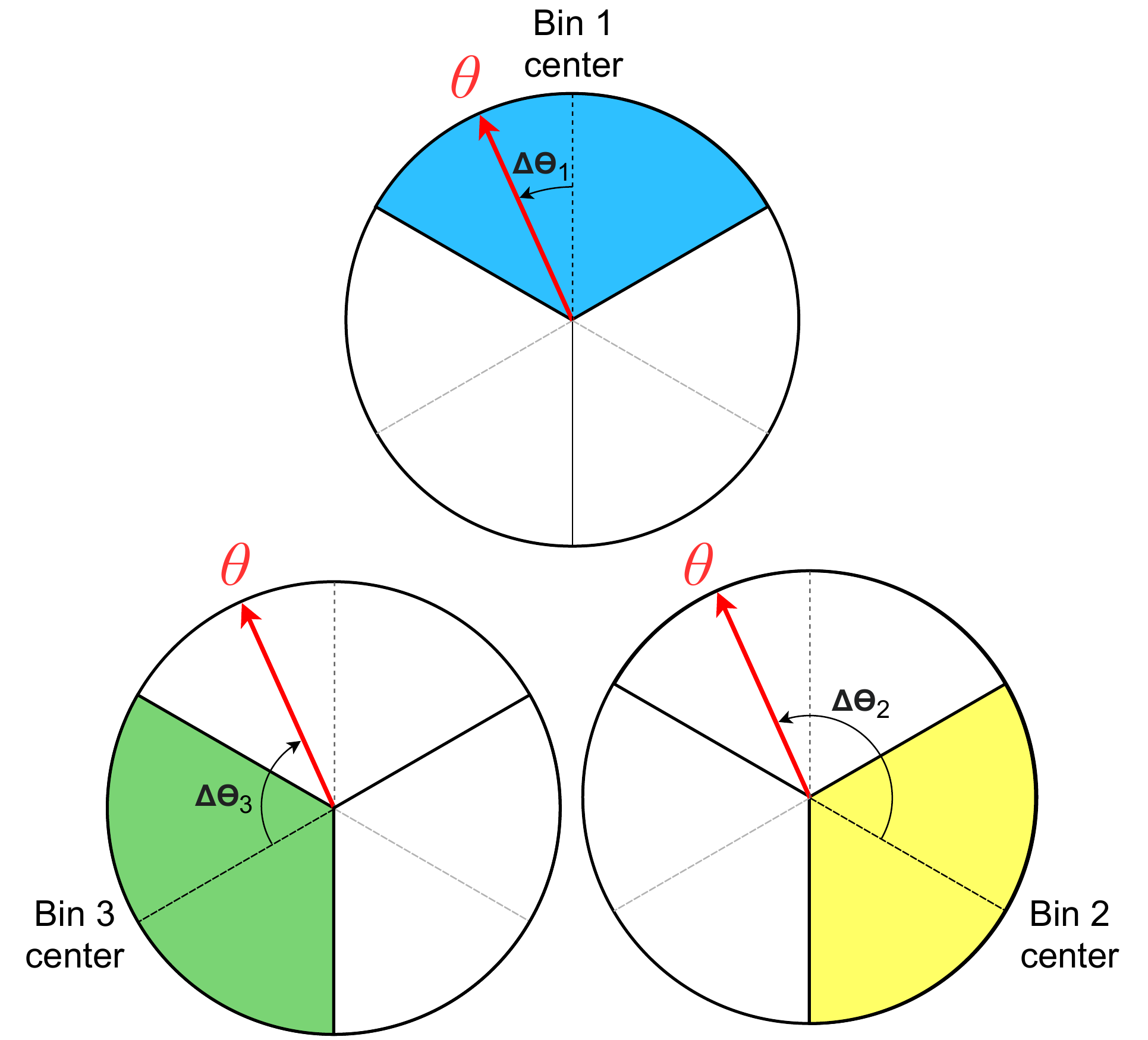}
    \caption{The Tricosine representation encodes the input angle with three cosine values. This allows for bin classification and regression within a continuous representation. This comes from the highest cosine value providing the bin center to create an average.}
    \label{fig:Tricosine}
\end{figure}

% triple sin() likely does not work because it is no longer a maximization problem, even though it instantly give which side of the center you are on. 
% SOOOO we don't need it because we have the other TWO cosine values to see which side of the bin we are on!

Unlike the previous methods, Tricosine aims to encode an angle in radians into the minimum number of bin affinity values that can unambiguously converted back into radians, i.e three. It utilizes three bins, each containing the cosine of the angular offset to each bin's center.

Converting back a singular angle displays the affinities' natural hierarchy. The highest affinity value first determines what bin the angle lies in. The inverse cosine of each affinity provides a relative position for our angle, where we utilize the other affinities' inverted values to sway our prediction, further supplying accuracy. This swaying is a simple mean of each inverted affinity value, relative to the classified bin. 

\section{Model and Training Setup}
% explain what this sections is about

\subsection{Dataset}
 As a widely used benchmark for many visual tasks including 2D Object detection, Bird's Eve View (BEV) tasks, depth estimation, and 3D object detection, the KITTI dataset \cite{kitti_geiger_2012} contains 7481 training images and 7513 testing images captured by the stereo camera and LiDAR sensors. Each instance is labeled with 2D and 3D bounding box location, 3D bounding box dimensions, orientation in $\alpha$ (alpha) and $r_y$ (rotation y). Other values are not used in our analysis. As an assumption of the KITTI dataset, vehicles only rotate with respect to the yaw axis, known as canonical orientation, so vehicle instances are always on a flat road instead of climbing a ramp. Scenes are restricted to highways and rural areas. 
 
\subsection{Data Preprocessing}
 In data prepossessing, we applied horizontal flipping image augmentation with the probability of 0.5 to the cropped image\cite{peng2021ocm3d}, and then converted the orientation labels to the different prediction representations. For example, before training the Voting Bin model, we convert the rotation y angles into four discrete bin affinity values.
 
 All instances are cropped with their 2D bounding box and resized to 224x224. For feature extraction, we use Xception Net\cite{chollet2017xception} as our backbone, achieving a good balance between model size and accuracy\cite{bianco2018benchmark}. In the final output layers, two fully connected layers are used, with 1024 and N-D output nodes respectively.
 
\subsection{Model Architecture}
% provided a figure to show the training architecture

\begin{figure}[H]
    \centering
    \includegraphics[width=120mm]{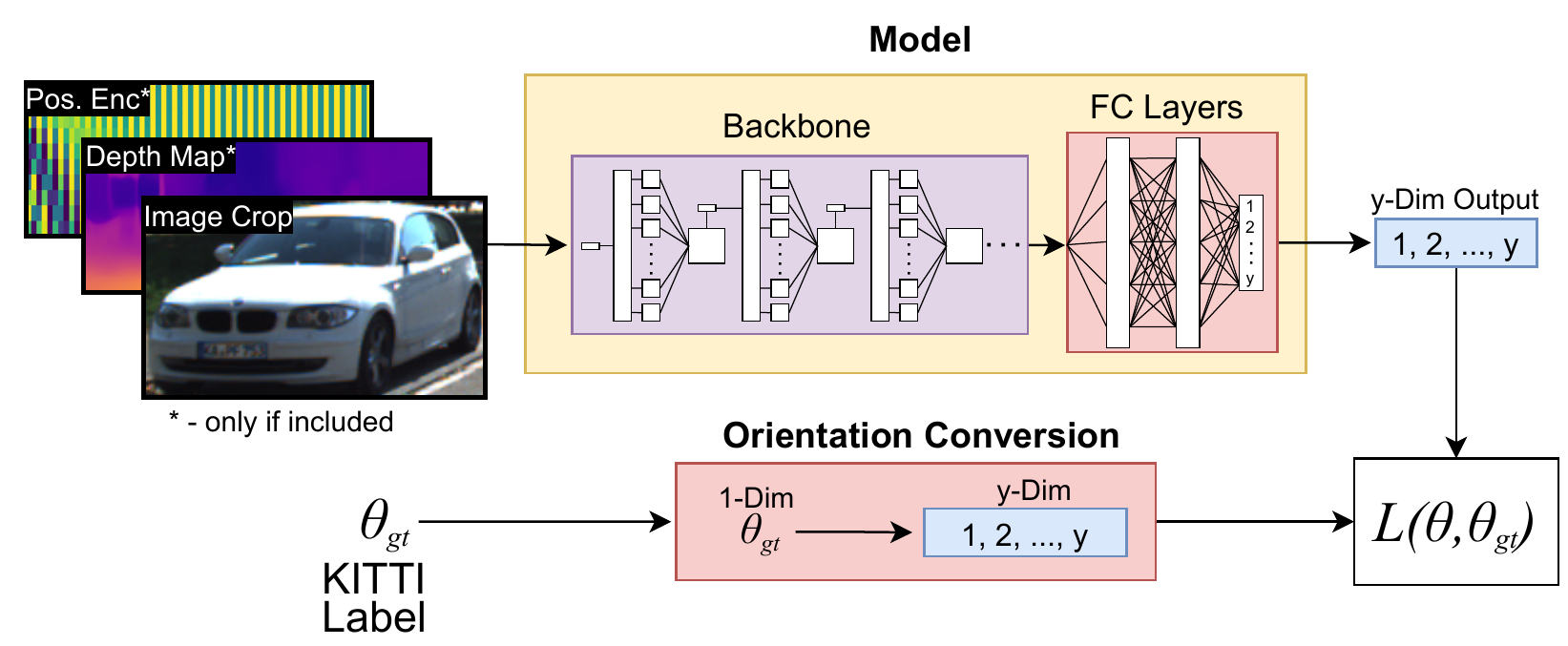}
    \caption{Illustration of Model and Training Process}
    \label{fig:model_diagram}
\end{figure}

Our contributions include comparison of these angular representations to monocular image inputs and the 'Car' class to remove potential geometric shortcuts gained from LiDar and stereo imagery and prevent a cost competition between classes \cite{voting_bin_zhao_2020} respectively. 
 
\subsection{Loss Functions}
% explain and list out all the loss function equations

We used simple mean squared error (L2 loss) as the loss function for all of the training experiments except for experiment E2. 
\begin{equation}\label{eq:l2loss}
    L(\hat{\theta}, \theta) = (\hat{\theta} - \theta)^2
\end{equation}
Here ${\hat{\theta}}$ is the prediction vector, and $\theta$ is the ground truth vector after converting to the corresponding angle representation.

Following \cite{hara2017designing}, we also consider an angular loss form to train on 1-D global rotation, or $r_y$. The angular loss function focuses only on the angular differences between angles, whereas the L2 loss computes the mean squared error on the scalar value differences for each component in the angle representation vector, which considers the norm of the vector representation and introduces the scalar \textit{discontinuity} problem (see section \ref{section:introduction}).

Angular loss function can be expressed as:
\begin{equation}\label{eq:angularloss}
L(\theta,\hat{\theta})=1-\cos (\theta)=1-\frac{\hat{\theta} \cdot \theta}{|\hat{\theta}
||\theta|}=1-\frac{x_{g} x+y_{g} y}{\sqrt{x^{2}+y^{2}}}
\end{equation}
Here $\hat{\theta} = [x, y]$ is the predicted vector, where $x = cos(\hat{\theta})$,  $y = sin(\hat{\theta})$, and $\theta = [x_g, y_g]$ is the ground truth vector, where $x_g = cos(\theta), y_g = sin(\theta)$.

\subsection{Accuracy Metric}
For all prediction methods, we compute the Orientation Similarity (OS) metric as provided by KITTI \cite{kitti_geiger_2012} without recall consideration. The cosine-like orientation similarity metrics computes the accuracy of predicted and ground truth angles in 1-D scalar representation, and maps the prediction and ground truth cosine distance to between [0, 1].

% why can't we use AOS

\begin{equation}
OS({\hat{\theta}}, {\theta}) = \frac{1}{N}\sum_{i=1}^{N} \frac{1+cos(\hat{\theta_{i}} - \theta_{i})}{2}
\end{equation}

\subsection{Training Setup}
We train our model using prediction methods, Global Rotation ($r_y$), Local Rotation ($\alpha$), Single Bin, Tricosine, Voting Bin, and Confidence Bins to predict one of two targets: global rotation ($r_y$) and location rotation ($\alpha$). Our models were first trained on a Nvidia V100 GPU for 1-D scalar based representation experiments and later on a Nvidia RTX3090 GPU for all other experiments. All models were trained with TensorFlow's default Adam optimizer \cite{kingma2014adam} (lr= 0.001, $\beta1$= 0.9, $\beta2$= 0.99, $\epsilon$= 1e-7), a batch size of 25, and 100 total epochs.

\section{Experiment Results Discussion}

In the section, we will provide a discussion on the key findings of our experiments. An overview of our experiment results is provided in table \ref{table:all_exp}.
    \begin{table}[h]
        \centering
        \begin{tabularx}{\textwidth} { 
            | 
            >{\centering\arraybackslash}X | 
            >{\centering\arraybackslash}X | 
            >{\centering\arraybackslash}X |
            >{\centering\arraybackslash}X |
            >{\centering\arraybackslash}X |}
            \hline
            Exp ID        & Prediction Method                  & Loss Functions           & Additional Inputs & Accuracy  (\%)    \\
            \hline\hline
            E1            & Global Rot              & L2 Loss           &-         &  90.490           \\
            E2            & Global Rot              & Angul. Loss       &-         &  89.052           \\
            E3            & Local Rot             & L2 Loss           &-         &  90.132           \\
            E4            & Single Bin              & L2 Loss           &-         &  \textbf{94.815}  \\
            E5            & Single Bin              & L2 Loss           &Pos Enc   &  94.277           \\
            E6            & Single Bin              & L2 Loss           &Dep Map   &  93.952           \\
            E7            & Voting Bins             & L2 Loss           &-         &  93.609           \\
            E8            & Tricosine               & L2 Loss           &-         &  94.249           \\
            E9            & Tricosine               & L2 Loss           &Pos Enc   &  94.351           \\
            E10           & Tricosine               & L2 Loss           &Dep Map   &  94.384           \\
            E11           & 2 Conf Bins             & L2 Loss           &-         &  83.304           \\
            E12           & 4 Conf Bins             & L2 Loss           &-         &  88.071           \\
            \hline
        \end{tabularx}
        \caption{Experiment result overview}
        \label{table:all_exp}
    \end{table}

    \subsection{1-D Global vs. 1-D Local Representation}
    % answer the research question, which is a better prediction target in scalar representation?
    % create a overlay line chart (val accuracy vs num epoch)
    % table of best val accuracy 
    % conclusion is no difference, use Rotation Y because it is easier to use in 2D and 3D visualizations
    Based on experiments E1 and E3, we compare two types of orientation representations in 1-D, global and local rotation representations, to answer: Which is a better prediction target in 1-D scalar orientation representation?

    \begin{figure}[h]
        \centering
        \includegraphics[width=100mm]{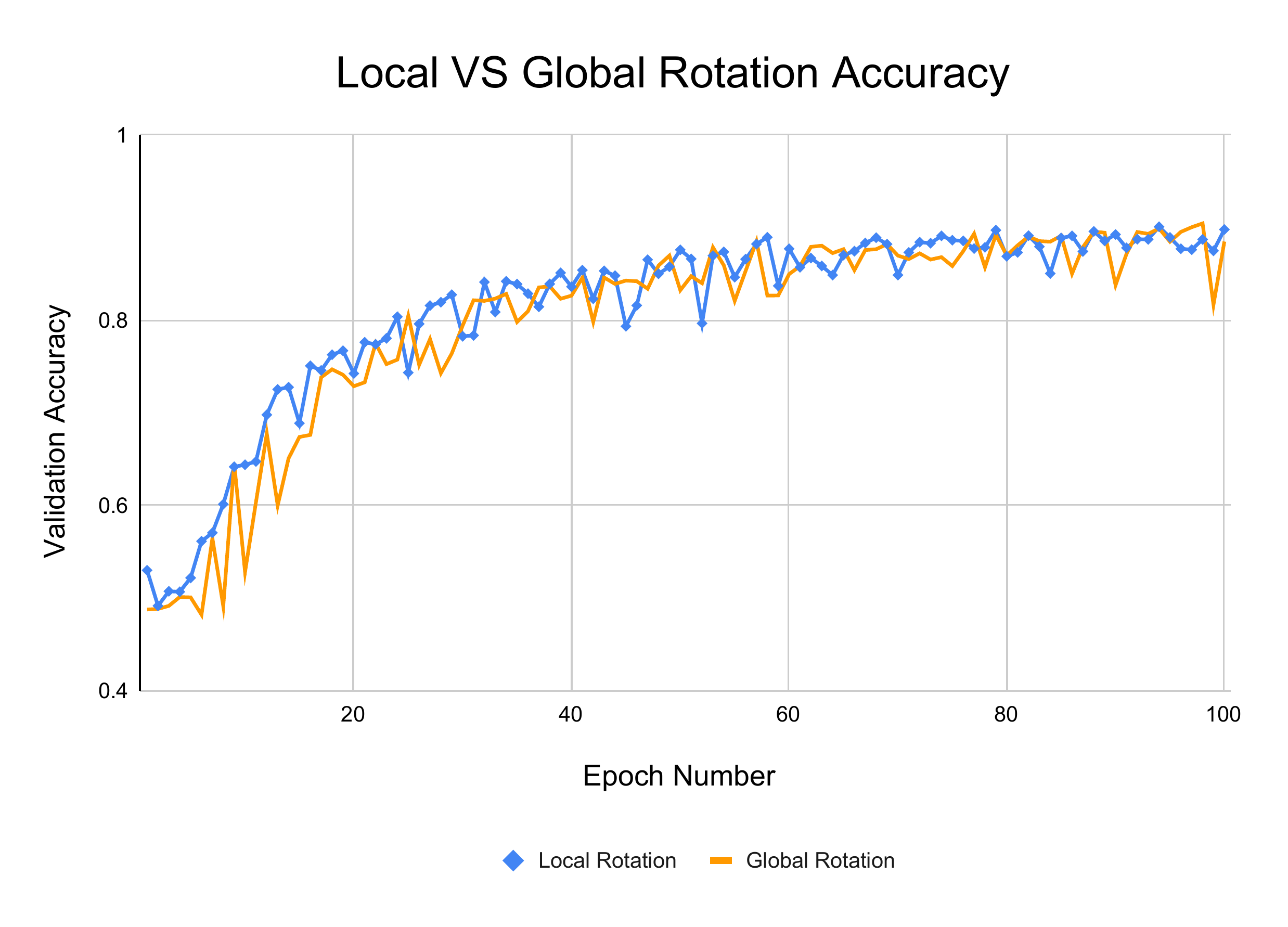}
        \caption{Max Validation accuracy of 1-D Global and local representation after 100 epochs of training.}
        \label{fig:alphaVSroty}
    \end{figure}

    \begin{table}[h]
        \centering
        \begin{tabularx}{\textwidth} { 
                    | 
                    >{\centering\arraybackslash}X | 
                    >{\centering\arraybackslash}X | 
                    >{\centering\arraybackslash}X | }
                    \hline
                    Target     & Accuracy (\%) \\ 
                    \hline\hline
                    Global Rotation      & 90.490   \\
                    Local Rotation   & 90.132   \\
                    \hline
        \end{tabularx}
        \caption{Validation accuracy of training with 1-D Global and 1-D Local representation across 100 epochs of training.}
        \label{table:alphaVSroty}
    \end{table}
    
    To answer this question, we initialized two models with the same backbone and output layers. We normalized target values to [-1, 1] from the dataset's range of [$-\pi$ and $\pi$]. The loss function used for both models during training was L2. 
    
    The training results, in table \ref{table:alphaVSroty}, show that the difference in the highest validation accuracy of 1-D Global and Local rotation is small, around 0.5\% after 100 epochs of training, which is within the margin of error between epochs. We conclude that there is either no difference or a slight disadvantage in using Local representation compared to using global representation.

    Thus, we speculate that the network is capable of detecting the global/egocentric orientation of an object using without needing the conversion into an angle based on the direction of camera observation.

\subsection{Angular Loss vs L2 loss}
% answer the research question, is it better to use angular loss?
% create a overlay line chart (val accuracy vs num epoch)
% table of best val accuracy
% conclusion is l2 is better
% explain why, probably due to the ambiguous convergence path

While angular loss removes the inherent radial length of single bin outputs by normalizing the bin's output, there are notable drawbacks that L2 does not encounter. For instance, angular loss's gradient encounters zeros within the prediction range [-2$\pi$, 2$\pi$], so an exact prediction of $\theta\pm\pi$ from ground truth will have no gradient to train the model's weights. Likewise, predictions near $\theta\pm\pi$ will suffer from diminished gradients and take a long time to converge. 

Since angular loss does not penalize the radial length of the output, we expected to have unbounded outputs paired with low loss. However, according to table \ref{table:angular_loss}, there was only little performance drop when using angular loss function. The batch-normalization layer \cite{ioffe2015batch} may explain this, as it clips the value range in case of output explosion and stabilize the training process. This can also be clamped at the model's output with a softmax layer, but this requires further testing. We do not believe it is necessary due to the low performance differences. 

% softmax to clamp down the range

\begin{figure}[h!]
        \centering
        \includegraphics[width=100mm]{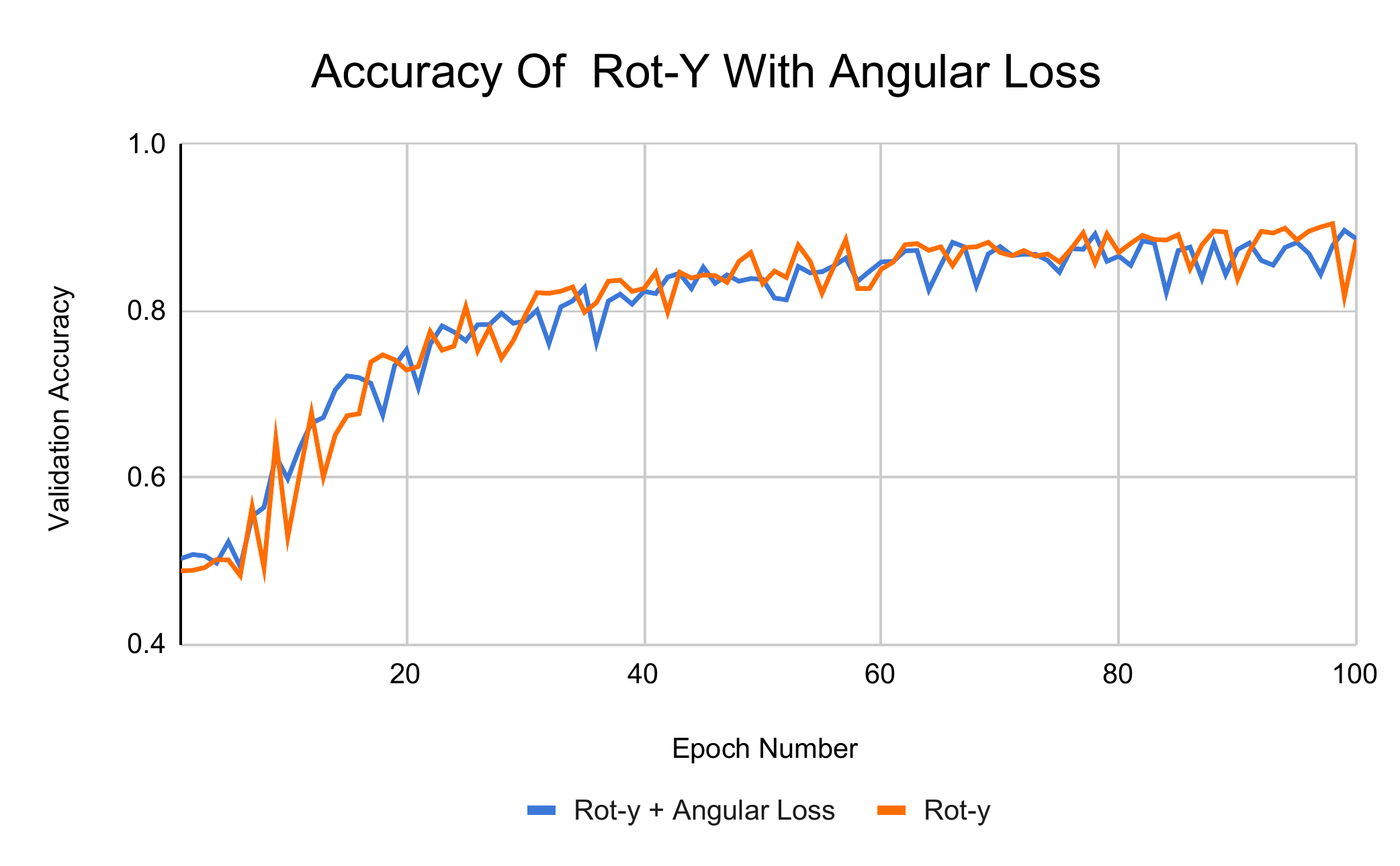}
        \caption{Validation Accuracy of prediction method Global Rotation with and without angular loss.}
        \label{fig:angular_loss}
    \end{figure}
    \begin{table}[h!]
        \centering
        \begin{tabularx}{\textwidth} { 
            | 
            >{\centering\arraybackslash}X | 
            >{\centering\arraybackslash}X | 
            >{\centering\arraybackslash}X | }
            \hline
            Method   & Loss Function    & Accuracy (\%) \\ 
            \hline\hline
            Global Rotation   & MSE    & 90.490 \\
             Global Rotation  & Angular Loss          & 89.052\\
            \hline
        \end{tabularx}
        \caption{Highest validation accuracy using two different loss function. Prediction target and prediction method are both Rotation Y}
        \label{table:angular_loss}
    \end{table}
 
\subsection{1-D vs. 2-D vs. N-D Bin Affinity Representation}
% answer the research question, is it better to use non-scalar continuous representation?
% create a overlay line chart (val accuracy vs num epoch)
% table of best val accuracy
% conclusion is 2D Cartesian is best
    We compare three types of orientation representations in 1-D, 2-D, 3-D, 4-D, and 8-D to answer the research question: Which vector dimensions of angle representation improve prediction accuracy?

    \begin{figure}[h]
        \centering
        \includegraphics[width=100mm]{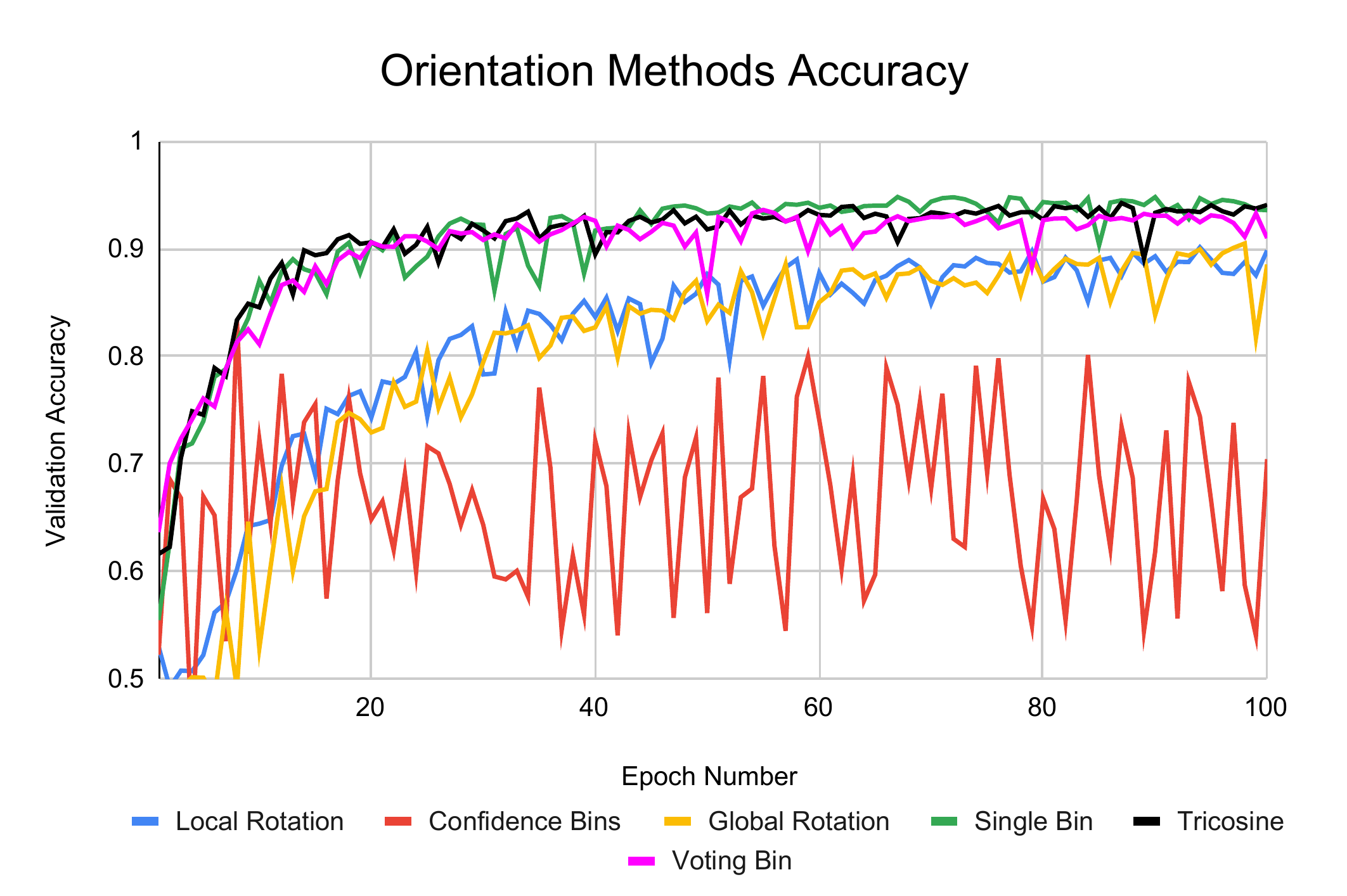}
        \caption{Max Validation accuracy of 1-D, 2-D, 3-D, 4-D, 8-D representation after 100 epochs of training.}
        \label{fig:orient_method_acc}
    \end{figure}

    \begin{table}[h]
        \centering
        \begin{tabularx}{\textwidth} { 
            | 
            >{\centering\arraybackslash}X | 
            >{\centering\arraybackslash}X | 
            >{\centering\arraybackslash}X | }
            \hline
            Target   & Method    & Accuracy (\%) \\ 
            \hline\hline
            1-D  & Global Rotation      & 90.490   \\
            1-D  & Local Rotation       & 90.132 \\
            2-D  & Single Bin           & 94.815   \\
            3-D  & Tricosine            & 94.249   \\
            4-D  & Confdence Bins       & 83.304  \\
            8-D  & VotingBin            & 93.609   \\

            \hline
        \end{tabularx}
        \caption{Highest validation accuracy on 5 prediction methods with prediction target of rot-y.}
        \label{table:orient_method_acc}
    \end{table}

Performance is measured by validation accuracy per epoch, which immediately divides into three groups with Single Bin, Tricosine, and Voting Bin performing the highest. Followed by the next  performance group, global and local representation, these two groups display expected performance differences, as 1-D angular representation contains the challenge of \textit{discontinuity} while Single Bin, Tricosine, and Voting Bin have compact, continuous representations. The third group contains Confidence Bin, which performs better than Multibin, but does not converge to the other N-D based methods.

\subsection{Positional Encoding}
% answer the research question, is it better to add positional information?
% create a overlay line chart (val accuracy vs num epoch)
% table of best val accuracy
% conclusion is no improvement, donno why
    \begin{figure}[!h]
    \begin{minipage}[!h]{0.49\textwidth}
        \includegraphics[width=60mm]{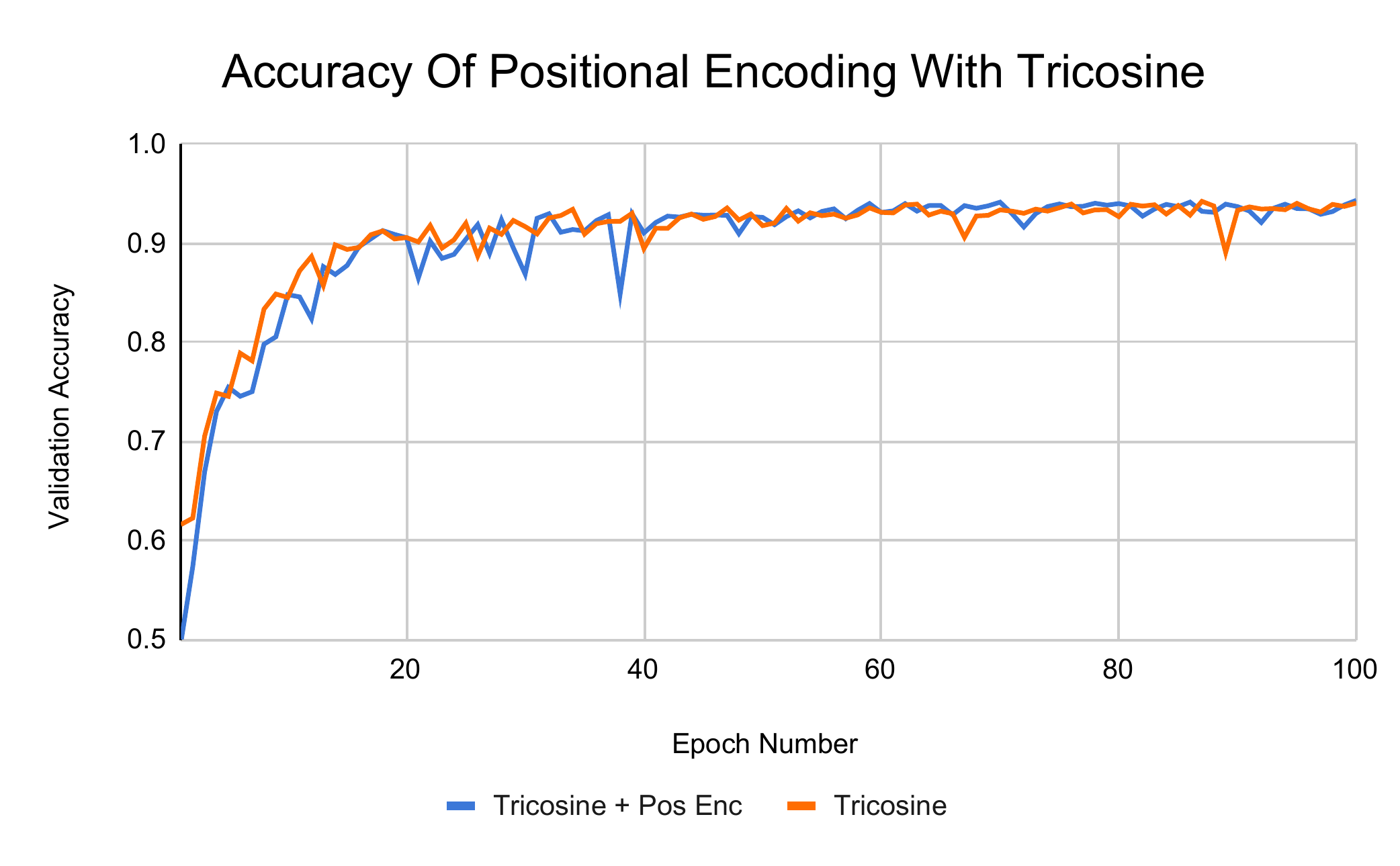}
    \end{minipage}
    \begin{minipage}[!h]{0.49\textwidth}
     \centering
        \includegraphics[width=60mm]{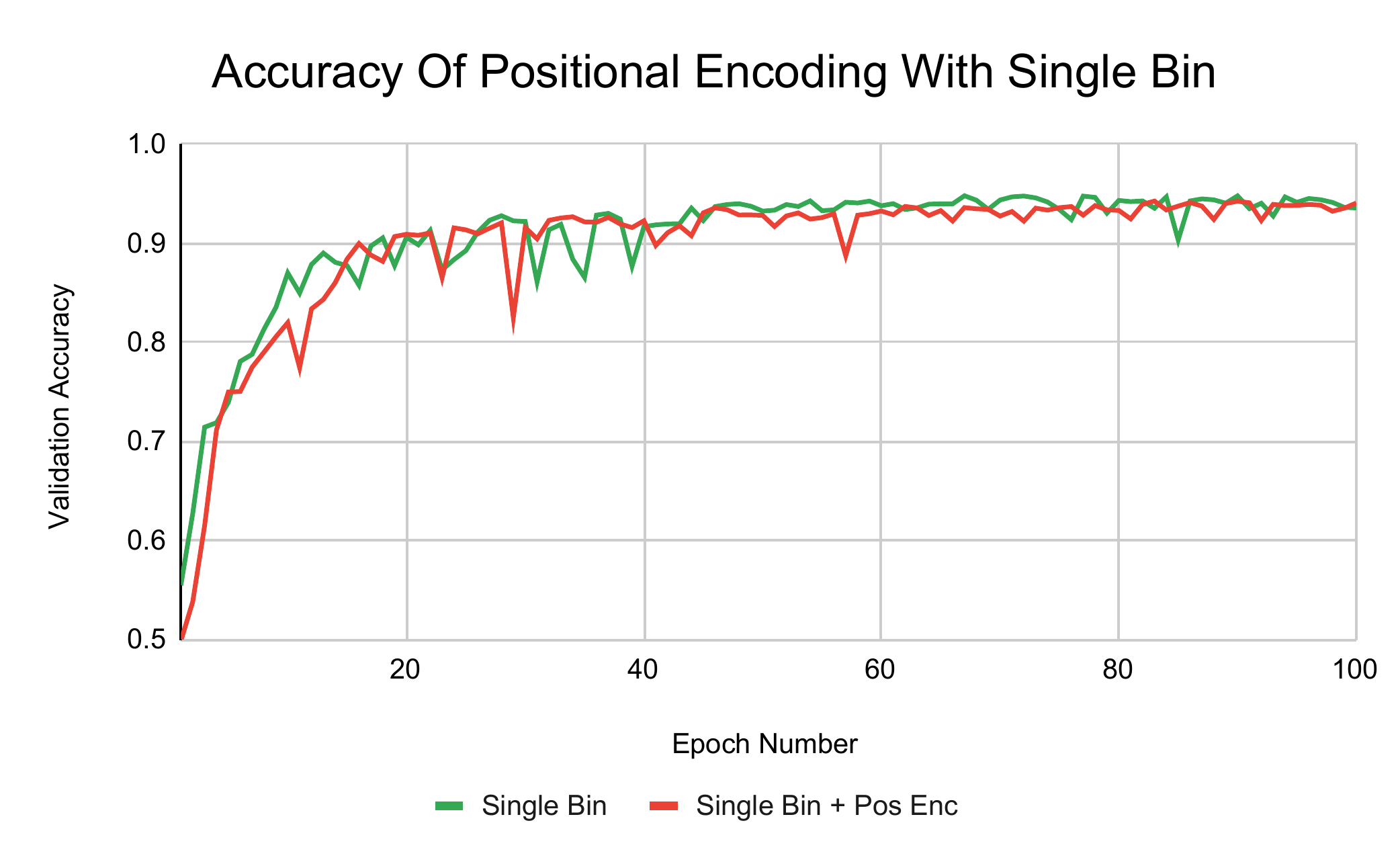}
    \end{minipage}
    \caption{Validation accuracy comparison between models with and without the positional encoding layer appended to input across 100 epochs.}
    \label{fig:pos_enc}
    \end{figure}
    
    \begin{table}[!h]
        \centering
        \begin{tabularx}{\textwidth} { 
            | 
            >{\centering\arraybackslash}X | 
            >{\centering\arraybackslash}X | 
            >{\centering\arraybackslash}X | }
            \hline
            Target  & Additions  & Accuracy \%\\ 
            \hline\hline
            Single Bin  &-        & 94.815   \\
            Single Bin  &PosEnc   & 94.277   \\
            Tricosine  &-        & 94.249   \\
            Tricosine  &PosEnc   & 94.351   \\
            \hline
        \end{tabularx}
        \caption{Highest validation accuracy comparison between models with and without positional encoding layer.}
        \label{tab:pos_enc}
    \end{table}

As only image crops are inputted to the network, we observe that crops' positions are ambiguous. Inspired by transformers \cite{vaswani2017attention}, we concatenate \cite{ke2020rethinking, li2021learnable, Gonzalez_2021_CVPR,dosovitskiy2020image,cheng2021sign,sun2021rethinking} a positional encoding channel to the RGB input and feed them as 4-channel images into the network. We hope to provide the crop's positional information to the model, and thus improve the $r_y$ accuracy. However, as shown in Table \ref{tab:pos_enc}, positional encoding provides little gain when predicting against $r_y$. We are not sure about the exact reason. Still, we speculate that the model either ignores the added channel or can estimate the position to some extent via the distortion\cite{sitzmann2018end} of the object within the crop.

\subsection{Depth Map + Image Crop vs. Image Crop Only}
% answer the research question, is it better to add depth information?
% create a overlay line chart (val accuracy vs num epoch)
% table of best val accuracy
% conclusion is no improvement, donno why
Similar to adding positional encoding to input, we want to test if the model can perform better if more depth information is provided. We generated depth maps for all RGB inputs with LapDepth model \cite{song2021lapdepth} in advance, and concatenated\cite{fu2020jl,Shigematsu_2017_ICCV, de2020infrared} them to the RGB images as input, which then is fed into the network. As listed in the table \ref{tab:depth}, we don't see any statistically significant improvement in accuracy. There is no solid explanation of it but we hypothesize that the model has the ability to estimate the depth of objects in the image. 

    \begin{figure}[!h]
    \begin{minipage}[!h]{0.49\textwidth}
        \includegraphics[width=60mm]{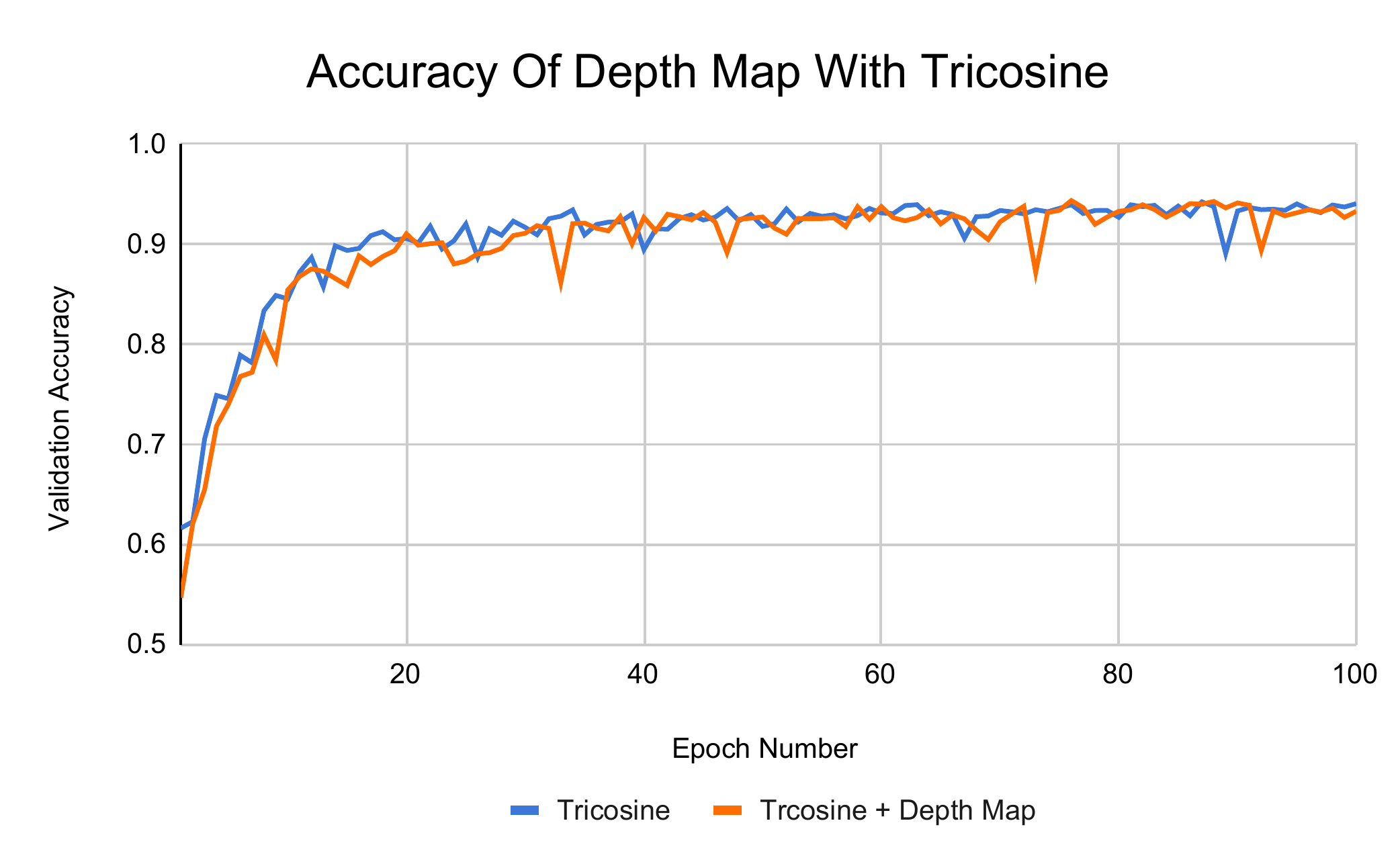}
    \end{minipage}
    \begin{minipage}[!h]{0.49\textwidth}
     \centering
        \includegraphics[width=60mm]{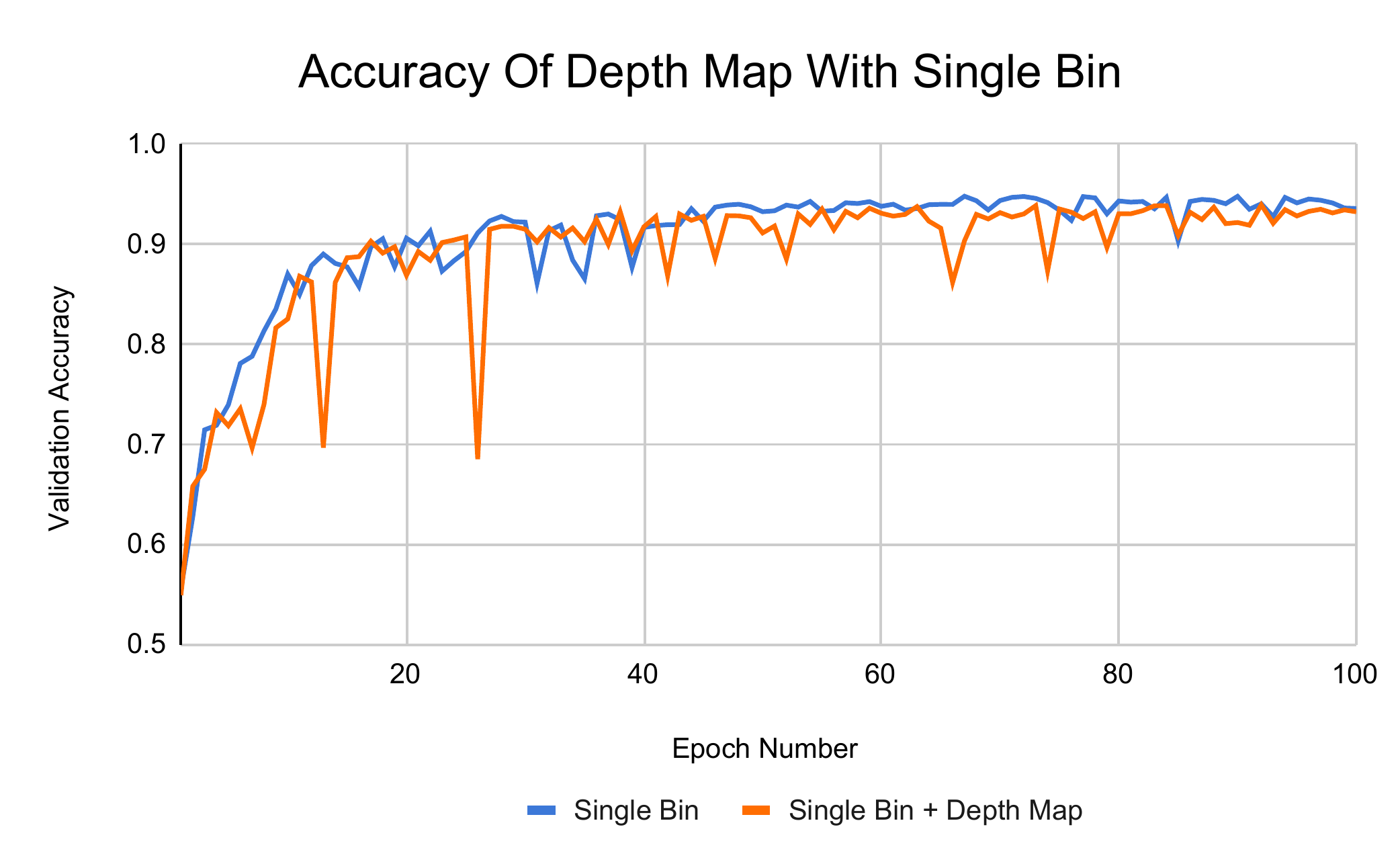}
    \end{minipage}
    \caption{Validation accuracy comparison between models with and without the positional encoding layer appended to input across 100 epochs.}
    \label{fig:pos_enc}
    \end{figure}
    
    \begin{table}[!h]
        \centering
        \begin{tabularx}{\textwidth} { 
            | 
            >{\centering\arraybackslash}X | 
            >{\centering\arraybackslash}X | 
            >{\centering\arraybackslash}X | }
            \hline
            Target  & Concatenations  & Accuracy \%\\ 
            \hline\hline
            Single Bin  &-        & 94.815   \\
            Single Bin  &depth map   & 93.952   \\
            Tricosine  &-        & 94.249   \\
            Tricosine  &depth map   & 94.384   \\
            \hline
        \end{tabularx}
        \caption{Highest validation accuracy comparison between models with and without depth map in input.}
        \label{tab:depth}
    \end{table}
    
% section for losses begins:
% put in the image of L2 vs angular loss vs 

\section{Conclusion}
% what research answer this paper answered (without citing the results), in English?
% single bin did best
% other inputs (position or depth) did not improve results (a surprising finding)
% Multibin/or more complicated representation caused bad results, probably because has a REAAAALLY unsmooth loss function due to overlap

% what are the shortcoming in our methods?  Has Your Research Left Some Unanswered Questions?
% we don't know if other inputs (LiDAR, multi-cams) would improve results, since we didn't test them.
% we don't know if this is also good for other axis, since we didn't test them.
% we don't know if there are other more complicated representations that could have performed better, since we didn't test them.
In this work, we performed extensive experiments on three types of orientation representations: 1-D scalar, 2-D Cartesian, and N-D bin affinity representations. We found that Single Bin with 2-D Cartesian representation achieved the highest accuracy. We also proposed and tested a novel way of predicting 3D orientation: Tricosine which achieves second-to-the-best validation accuracy. Although multiple existing works implement the Multibin representation, we found that it did not perform well, potentially because of its unsmooth loss function. Confidence Bins has much greater performance than Multibin as it has a smooth loss function, but it did not perform as well as Tricosine or Single Bin with 2-D Cartesian representation. 

% CITE THIS, single bin?
These results concretely confirm other works' conclusion regarding the problem of non-continuity in scalar representation, as the representations with continuous and unambiguous representations such as Single Bin, Tricosine, and Voting Bin all achieved the greater performance than scalar representation.

Additionally, we found that using local rotation targets did not outperform global rotation targets. This was surprising as we expected global rotation to perform poorly due to the major visual differences between crops of constant global rotation. We also tested if assisting the image crops with the location of the crop within the image can have improved performance, by adding positional encoding. However, our results from adding positional encoding to predict global rotation target gained little performance, and at times destabilized validation accuracy across training epochs. 

Our work was limited in using only monocular object image crops as inputs. Perhaps lidar-based and multi-camera inputs may provide further improvements to the orientation estimation tasks. We look to future work to see if our methods also apply to full 3-DoF rotation and other orientation representation methods.

\clearpage
\newpage
\section{Appendix}

\clearpage
\newpage
\printbibliography

\end{document}